\documentclass{article}

\usepackage{microtype}
\usepackage{graphicx}
\usepackage{booktabs} %
\usepackage{subfig}
\usepackage{multirow}

\usepackage{hyperref}

\usepackage{arxiv}
\usepackage{natbib}
\usepackage{wrapfig}
\usepackage{xcolor}
\definecolor{mydarkblue}{rgb}{0,0.08,0.45}
\hypersetup{
  colorlinks=true,
  linkcolor=mydarkblue,
  citecolor=mydarkblue,
  filecolor=mydarkblue,
  urlcolor=mydarkblue}

\usepackage[utf8]{inputenc} %
\usepackage[T1]{fontenc}    %
\usepackage{amsmath}
\usepackage{amssymb}
\usepackage{amsfonts}
\usepackage{nicefrac}
\usepackage{siunitx}

\def\X{{\mathbf X}}

\newcommand{\R}{\mathbb{R}}

\newcommand{\voireq}[1]{Equation~(\ref{#1})}
\newcommand{\voirsec}[1]{Section~\ref{#1}}
\newcommand{\voirfig}[1]{Figure~\ref{#1}}
\newcommand{\voirtbl}[1]{Table~\ref{#1}}
\newcommand{\bX}{\mathbf{X}}

\begin{document}

\title{Word-level Speech Recognition\\ with a Letter to Word Encoder}

\renewcommand{\cite}[1]{\citep{#1}}
\definecolor{mydarkblue}{rgb}{0,0.08,0.45}
\hypersetup{ %
  colorlinks=true,
  linkcolor=mydarkblue,
  citecolor=mydarkblue,
  filecolor=mydarkblue,
  urlcolor=mydarkblue}

\setcitestyle{authoryear,round,citesep={;},aysep={,},yysep={;}}

\author{
  Ronan Collobert
  \quad\enskip
  Awni Hannun
  \quad\enskip
  Gabriel Synnaeve \\
  Facebook AI Research\\
  \texttt{\{locronan,awni,gab\}@fb.com} \\
}

\maketitle

\begin{abstract}
We propose a direct-to-word sequence model which uses a word network to learn
word embeddings from letters. The word network can be integrated seamlessly
with arbitrary sequence models including Connectionist Temporal
Classification and encoder-decoder models with attention. We show our
direct-to-word model can achieve word error rate gains over sub-word level
models for speech recognition. We also show that our direct-to-word
approach retains the ability to predict words not seen at training time
without any retraining. Finally, we demonstrate that a word-level model can
use a larger stride than a sub-word level model while maintaining accuracy.
This makes the model more efficient both for training and inference.
\end{abstract}

\section{Introduction}
\label{sec-intro}

Predicting words directly has the potential to allow for more accurate, more
efficient and simpler end-to-end automatic speech recognition (ASR) systems.
For example, outputting words enables a word-level learning objective which
more directly optimizes the quality of the transcription. While phonemes last
50-150 milliseconds and graphemes can be shorter or even silent, words are on
average much longer. This allows the acoustic model to capture meaning more
holistically, operating at a lower frequency with a larger context. To operate
at a lower frequency the acoustic model takes larger strides (e.g. sub-samples
more), which contributes to a much faster transcription speed and a lower
memory footprint.

One major hurdle for direct-to-word approaches is transcribing words not found
in the training vocabulary. Unlike phone-based or letter-based systems, the
output lexicon in a word-based model is bounded by the words observed in the
training transcriptions. Here, we propose an end-to-end model which outputs
words directly, yet is still able to dynamically modify the lexicon with words
not seen at training time. The method consists in jointly training an acoustic
model which outputs word embeddings with a sub-word model (e.g. graphemes or
word pieces) which also outputs word embeddings. The transcription of a new
word is generated by inputting the individual tokens of the word into the
sub-word model to obtain an embedding. We then match this embedding to the
output of the acoustic model. For instance, assume the word ``caterpillar'' was
not seen at training time, but ``cat'' and ``pillar'' were.  If we input
``caterpillar'' into the word embedding model it should yield an embedding
close to the embedding of the acoustic model when observing speech with the
words ``cat'' and ``pillar''.

Another hurdle for direct-to-word approaches is the need for massive datasets
since some words may be seen only a few times in a data set with even 1,000
hours of speech. For example, \citet{soltau2016neural} demonstrate competitive
word-based speech recognition but require 100,000 hours of captioned video to
do so. Our approach circumvents this issue by learning an embedding from the
sub-word tokens of the word. By jointly optimizing the word model and the
acoustic model, we gain the added advantage of learning acoustically relevant
word embeddings. In order to efficiently optimize both the acoustic and word
models simultaneously, we use a simple sampling based approach to
approximate the normalization term over the full vocabulary. We also show how
to efficiently integrate the word model with a beam search decoder.

We validate our method on two commonly used models for end-to-end speech
recognition. The first is a Connectionist Temporal Classification (CTC)
model~\cite{graves2006} and the second is a sequence-to-sequence model with
attention (seq2seq)~\cite{bahdanau2014neural, cho2014learning,
sutskever2014sequence}. We show competitive performance with both approaches
and demonstrate the advantage of predicting words directly, especially when
decoding without the use of an external language model. We also validate that
our method enables the use of a dynamic lexicon and can predict words not seen
at training time more accurately than a word piece baseline. Finally, we show
that with our word level model, we are able to surpass state-of-the-art word
error rate (WER) for end-to-end models on a low resource speech recognition
benchmark.

\section{Related Work}

Our work builds on a large body of research in end-to-end sequence models in
general and also in speech recognition~\citep{amodei2016,
bahdanau2016icassp, chan2016listen, collobert2016}. The direct-to-word approach
can be used with structured loss functions including Connectionist Temporal
Classification (CTC)~\cite{graves2006} and the Auto Segmentation Criterion
(ASG)~\cite{collobert2016} or less structured sequence-to-sequence models with
attention~\cite{bahdanau2014neural}.

Unlike lexicon-free approaches~\cite{likhomanenko2019, maas2015lexicon,
zeyer2018improved} our model requires a lexicon. However, the lexicon is
adaptable allowing for a different lexicon to be used at inference than that
used during training. Traditional speech recognizers also have adaptable
lexicons, which they achieve by predicting phonemes and requiring a model for
grapheme-to-phoneme conversion~\cite{bisani2008joint, rao2015grapheme}. Direct
to grapheme speech recognition~\cite{kanthak2002context, killer2003grapheme}
circumvents the need for the grapheme-to-phoneme model but suffers from other
drawbacks including (1) inefficiency due to a high-frame rate and (2)
optimizing the letter error rate instead of the word error rate. In contrast,
we show that it is possible to predict words directly while retaining an
adaptable lexicon.

Prior work has also attempted direct-to-word speech
recognition~\cite{audhkhasi2017direct, li2018advancing, soltau2016neural}.
These approaches require massive data sets to work well~\cite{soltau2016neural}
and do not have adaptable lexicons. Similar to our work is that
of~\citet{bengio2014word}, who learn to predict word embeddings by using a
sub-word level embedding model. However, their model is not end-to-end in that
it requires an aligned training dataset and only uses the word model for
lattice re-scoring. They also use a triplet ranking loss to learn the word
embeddings, whereas our word model integrates seamlessly with common loss
functions such as CTC or categorical cross entropy.  Building on this work,
\citet{settle2019words} keep the triplet loss but update the model to an
end-to-end architecture. Their lexicon was, however, limited to up to 20k
words.

One benefit of the direct-to-word approach is that optimizing a word-level loss
can lead to a more direct improvement in WER. Prior work has attempted to build
structured loss functions which optimize a higher-level error metric.
Discriminative loss functions like Minimum Bayes Risk training can be used to
optimize the WER directly but are not end-to-end~\cite{gibson2006hypothesis}.
On the other hand, end-to-end approaches are often
non-differentiable~\cite{graves2014towards, prabhavalkar2018minimum}, or can be
quite complex and difficult to implement~\cite{collobert2019}. In contrast, the
word model we propose is easy to use with existing end-to-end models.

Similar approaches to building word representations from characters have been
studied for tasks in language processing~\cite{bojanowski2017enriching,
kim2016character, labeau2017character, ling2015finding}.
\citet{santos2014learning} apply a convolutional model to construct word
embeddings from characters. Our work differs from this work in that (1) we show
how to integrate word-level embeddings with structured loss functions like CTC
that are commonly used in speech recognition and (2) we use a sampling approach
to efficiently compute the lexicon-level normalization term when training the
model.  \citet{ling2015character} propose a hierarchical character-to-word
based sequence-to-sequence model for machine translation.  This approach
generates new words character-by-character and hence is more computationally expensive and not
easily generalizable for use with structured loss functions like CTC.

\section{Model}

\begin{figure*}[t]
\centering
\begin{minipage}{.48\textwidth}
    \centering
    \includegraphics[width=\linewidth]{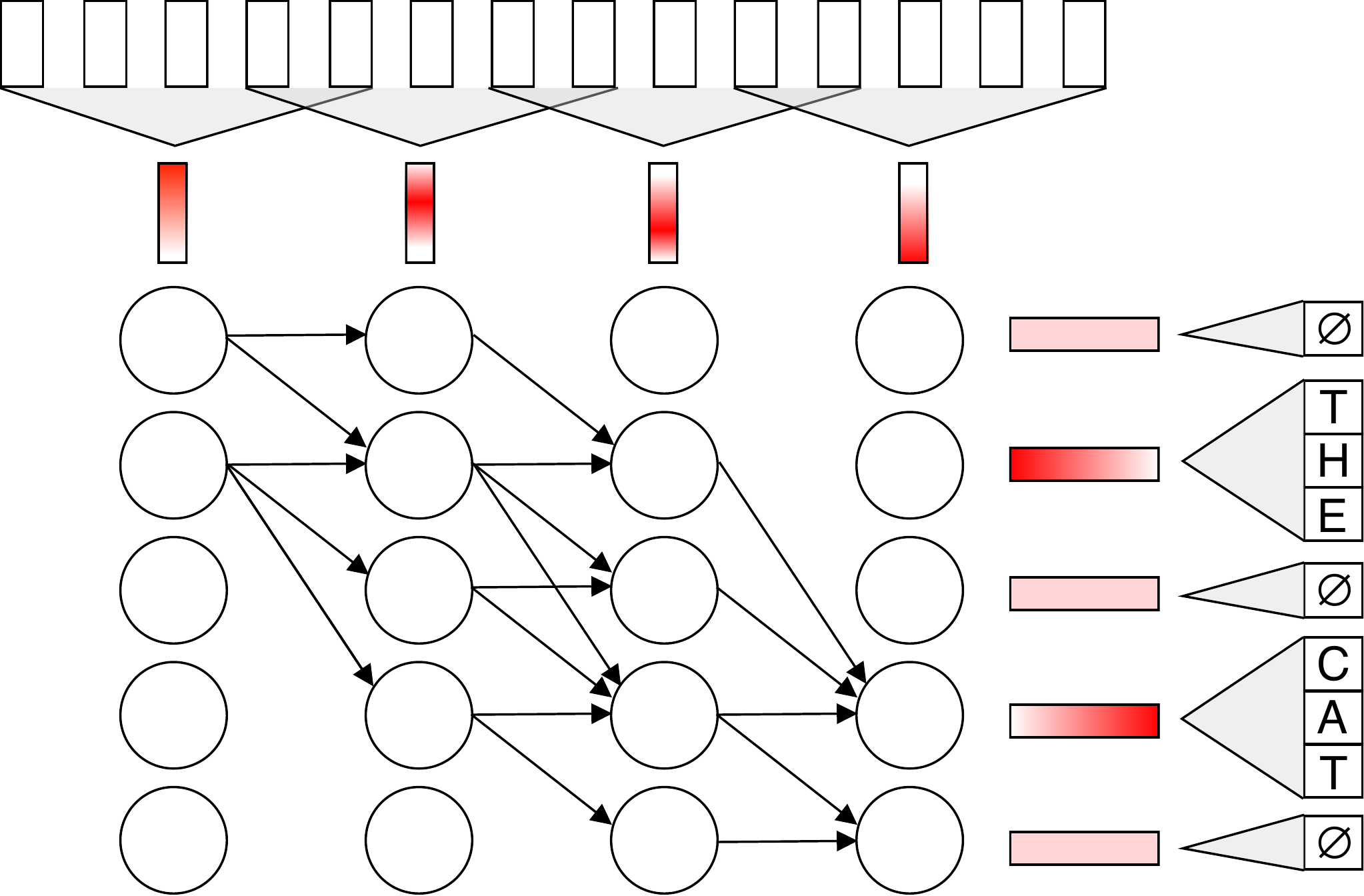}
    \caption{A CTC trained acoustic model combined with the letter-based word
    model. The $\varnothing$ denotes \textsc{blank}.}
    \label{fig:ctc_words}
\end{minipage}
\hspace{2mm}
\begin{minipage}{.49\textwidth}
    \centering
    \includegraphics[width=\linewidth]{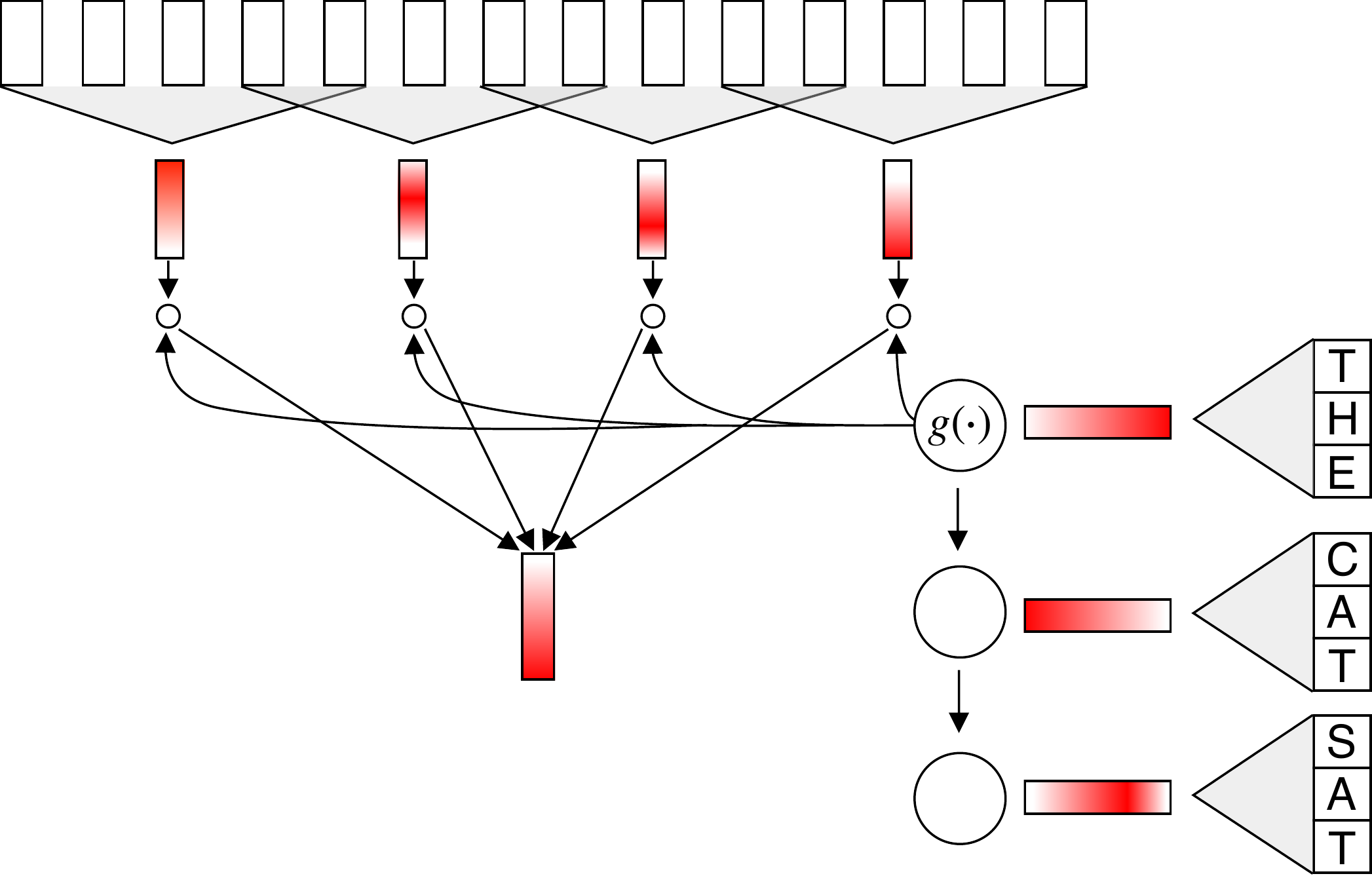}
    \caption{A seq2seq model combined with the letter-based word model. The
    function $g(\cdot)$ denotes the decoder Transformer.}
\label{fig:s2s_words}
\end{minipage}
\hspace{3mm}
\end{figure*}

We consider a standard speech recognition setting, where we are given
audio-sentence pairs $(\mathbf{X},\, \mathbf{Y})$ at training time.
$\mathbf{X}=\{\mathbf{x}_1,\, \dots, \mathbf{x}_T\}$ is a sequence of acoustic
frames (e.g. log-mel filterbanks), and $\mathbf{Y} = \{y_1,\,\dots y_N\}$ is a
sequence of words with each $y_i$ in the lexicon ${\cal D}$. We are interested
in models mapping $\mathbf{X}$ to $\mathbf{Y}$ in an end-to-end manner. We
consider two approaches: (1)~a structured-output learning approach, leveraging
a sequence-level criterion like CTC, and (2)~a sequence-to-sequence (seq2seq)
approach, leveraging an encoder-decoder model with attention.
Figures~\ref{fig:ctc_words} and~\ref{fig:s2s_words} give a high-level overview
of our proposed approach combined with CTC and seq2seq models respectively.

As mentioned in \voirsec{sec-intro}, word-level end-to-end approaches face
generalization issues regarding rare words, as well as challenges in
addressing out-of-vocabulary words (which commonly occur at inference time in
ASR). To cope with these issues, we consider an acoustic model
$f^{am}(\mathbf{W} |\mathbf{X})$ which relies on \emph{word
embeddings} to represent any word $y\in {\cal D}$ in the lexicon by a
$d$-dimensional vector $\mathbf{W}_y \in \R^d$. As the acoustic model relies
strictly on word embeddings to represent words, the model may operate on a
different lexicon ${\cal D'}$ (say at inference), assuming one can provide
corresponding word embeddings $\mathbf{W}'$.

Word embeddings $\mathbf{W}$ are computed with a specific network architecture
$f^{wd}(\cdot)$, which operates over sub-word units (in our case letters).
The acoustic model $f^{am}(\cdot)$ and word model $f^{wd}(\cdot)$ are trained
jointly, forcing both models to operate in the same acoustically-meaningful
$d$-dimensional embedding space. In the following, we detail the word model, as
well as the CTC and seq2seq combined acoustic and word model approaches.

\subsection{Letter-based Word Modeling}
\label{sec-wordembed}
\begin{wrapfigure}[24]{r}{0.25\textwidth}
\ifdim\pagetotal=0pt \else\vspace{-2\intextsep}\fi
\includegraphics[width=0.25\textwidth]{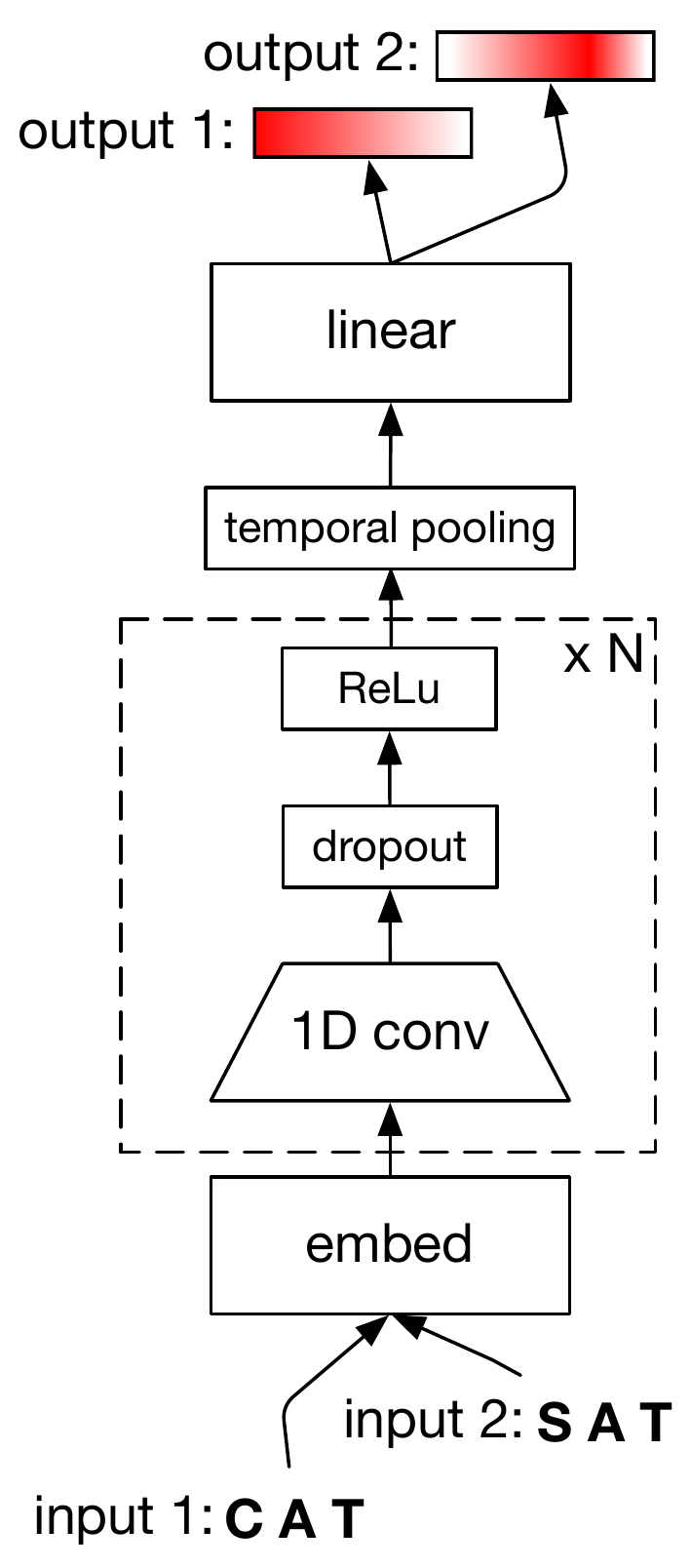}
\caption{
\label{fig:wordnet}
Architecture of the letter to word embedding model.
}
\end{wrapfigure}

The word model $f^{wd}(\mathbf{\sigma}(y))$ maps any word $y \in {\cal D}$ into
a $d$-dimensional space. As shown in \voirfig{fig:wordnet}, it takes as input
the sequence of letters $\mathbf{\sigma}(y)$ of the word, and performs a series
of 1D-convolutions interleaved with ReLU non-linearities. For our word models,
we use $N\!=\!3$ layers of convolution, the first at a stride of 1 and the
second and third at a stride of 2. Each output of the last convolution depends
on a fixed number ($n$) of input characters (as determined by the strides and
kernel sizes) which can be seen as an $n$-gram embedding for each input. A
fixed-sized embedding for the full word is obtained by a 1D max-pooling
aggregation over these n-gram representations, followed by a linear layer.

One can efficiently compute the embedding of all words in the dictionary
by batching several words together. When batching, an extra \textsc{<pad>}
character is used to pad all words to the same maximum word length.
Moreover, both CTC and seq2seq-based acoustic modeling rely on special word
tokens (like \textsc{blank} for CTC or the end-of-sentence token
\textsc{eos} for seq2seq). Embeddings for these special words are also
obtained with the same approach, using special \textsc{<blank>} and
\textsc{<eos>} tokens as input to the word model.

\subsection{Transformer-based Acoustic Modeling}
\label{sec-transformer}

Our Transformer-based acoustic models, $f^{am}(\mathbf{X})$, take as input a
sequence of acoustic frames $\mathbf{X}$ and output $d$-dimensional embeddings.
The architecture begins with a small front-end with $6$ layers of 1D
convolutions each of kernel width $3$ and respective input and output sizes
$(80,2048)$, $(1024,2048)$, $(1024,2048)$, $(1024,2048)$, $(1024,2048)$,
$(1024,1536)$. Each convolution is followed by a GLU activation
function~\cite{dauphin2017} and every other convolution strides by 2. The
output of the front-end is thus sub-sampled by a factor of $8$. After the
front-end, there are $L_e = 36$ Transformer blocks, and each block has $4$
self-attention heads with dimension $768$, followed by a feed forward network
(FFN), one hidden layer of dimension $3072$, and a ReLU non-linearity.

The output of the front-end is represented by the matrix $\mathbf{H^0} \in
\mathbb{R}^{m \times T}$ where $T$ in this case is the reduced number of frames
from sub-sampling in the convolution. Following the original architecture of
\citet{vaswani2017attention}, we have:
\begin{align*}
    \mathbf{Z}^{i} & = \textsc{Norm}(\textsc{SelfAttention}(\mathbf{H}^{i-1}) + \mathbf{H}^{i-1}), \\
    \mathbf{H}^{i} & = \textsc{Norm}(\textsc{FFN}(\mathbf{Z}^i) + \mathbf{Z}^i),
\end{align*}
where $\mathbf{Z}$ is the output of the self-attention layer, with a skip
connection, and $\mathbf{H}$ is the output of the FFN layer, with a skip
connection. Also following the original architecture, we use layer
normalization~\cite{ba2016layer} as the $\textsc{norm}(\cdot)$, and define
self-attention as:
\begin{align*}
    \textsc{SoftMax}(\frac{1}{\sqrt{h}} (\mathbf{A}_K\mathbf{H})^\top
(\mathbf{A}_Q\mathbf{H})_t)\mathbf{A}_V\mathbf{H}
\end{align*}
where $\mathbf{A}_K$, $\mathbf{A}_V$ and $\mathbf{A}_Q$ are key, value and
query projection matrices, respectively, and $h$ is the inner dimension of the
projected hidden states.

For CTC-trained models, the output of the Transformer blocks, $\mathbf{H}^{L_e}$, is
followed by a linear layer projecting to the number of output classes.
For all of the layers in the encoder (and the decoder when present), we use
dropout on the self-attention and layer drop following \citet{fan2019reducing}.

\subsection{CTC-based Acoustic Modeling}

Given an input sequence $\mathbf{X}$, the acoustic model is a Transformer-based
architecture (\voirsec{sec-transformer}), which outputs a sequence of embeddings:
\begin{equation}
  \label{eq-am-embed}
  f^{am}_t(\mathbf{X}) \in \R^d \quad\quad 1 \leq t \leq T\,.
\end{equation}
The log-probability $P_t(y|\mathbf{X}, \mathbf{W})$ of a word $y$ at each
time step $t$ is then obtained by performing a dot-product with the word
embedding of the word model from \voirsec{sec-wordembed}, followed by a log-softmax operation
over possible words in the lexicon ${\cal D}$.
\begin{equation}
  \label{eq-wordscore}
    \log P_t(y|\mathbf{X}, \mathbf{W}) = \mathbf{W}_y \cdot f^{am}_t(\mathbf{X}) - \log \sum_{y' \in {\cal D}} e^{\mathbf{W}_{y'} \cdot f^{am}_t(\mathbf{X})}\,.
\end{equation}
CTC~\cite{graves2006} is a structured-learning approach, which learns to align a sequence of
labels $\mathbf{Y}=\{y_1,\,\dots\,,y_N\}$ to an input sequence
of size $T$ (in our case the embeddings of~\voireq{eq-am-embed}). An alignment
$\pi=\{\pi_1,\,\dots,\,\pi_T\} \in {\cal D}^T$ over $T$
frames is valid for the label sequence $\mathbf{Y}$ if it maps
to $\mathbf{Y}$ after removing identical consecutive
$\pi_t$. For example, with $T=4$, $N=3$, ${\cal D} = \{a,\,b,\,c\}$ and a
label sequence ``cab'', valid alignments would be ``ccab'', ``caab'' and
``cabb''. In CTC we also have a \textsc{blank} label, which
allows the model to output nothing at a given time step,
and also provides a way to separate actual label repetitions in
$\mathbf{Y}$.

With CTC, we maximize the log-probability $\log P(\mathbf{Y}|\mathbf{X},
\mathbf{W})$, computed by marginalizing over all valid alignments:
\begin{equation}
  \label{eq-ctc}
  \log P(\mathbf{Y}|\mathbf{X}, \mathbf{W}) = \log \sum_{\mathbf{\pi} \in {\cal A}_{\mathbf{Y}}^T} e^{\sum_{t=1}^T \log P_t(\pi_t|\mathbf{X},\, \mathbf{W})}\,,
\end{equation}
where ${\cal A}_{\mathbf{Y}}^T$ is the set of valid alignments of length $T$
for $\mathbf{Y}$. The log-probability of~\voireq{eq-ctc} can be efficiently
computed with dynamic programming.

It is worth mentioning that the output sequence length $T$
in~\voireq{eq-am-embed} depends on the size of the input sequence $\mathbf{X}$,
as well as the amount of padding and the stride of the acoustic model
architecture. Successful applications of CTC are only possible if $T\geq N$
(where $N$ is the size of the label sequence $\mathbf{Y}$). When using sub-word
units as labels (like characters), using large strides in the acoustic model
may not be possible. In contrast, since we are using words as labels, $N$ is
much smaller than it would be with sub-word units, and we can afford
architectures with larger strides -- leading to faster training and inference.

\subsection{Seq2Seq-based Acoustic Modeling}
\label{sec-s2s}

Given an input $\bX$, the seq2seq model outputs an embedding for
each possible token in the output:
\begin{equation}
  f^{am}_n(\mathbf{X}) \in \R^d \quad\quad 1 \leq n \leq N\,.
\end{equation}
The log probability of a word $y$ at step $n$ is obtained
with~\voireq{eq-wordscore}. In our case, the embeddings $f_n^{am}(\X)$ are
computed by first \emph{encoding} the input $\bX$ with a Transformer encoder,
and then \emph{decoding} the resultant hidden states with a Transformer decoder
equipped with an attention mechanism. The encoder is given by:
\begin{equation}
    \begin{bmatrix} \mathbf{K}\\ \mathbf{V} \end{bmatrix} = \mathbf{H}^{L_e} = \text{encode}(\bX)
\end{equation}
where $\mathbf{K} = \{\mathbf{K}_1, \ldots, \mathbf{K}_T\}$ are the keys and
$\mathbf{V} = \{\mathbf{V}_1 \ldots, \mathbf{V}_T\}$ are the values. The decoder
is given by
\begin{align}
    \mathbf{Q}_n &= g(\mathbf{W}_{y_{<n}}, \mathbf{Q}_{<n}) \label{eq:attend} \\
    f_n^{am}(\bX) &= \mathbf{V}_n \cdot \textsc{softmax}\left(\frac{1}{\sqrt{d}} \mathbf{K}_n^\top \mathbf{Q}_n \right)
\end{align}
where $g(\cdot)$ is a decoder consisting of a stack of 6 Transformer blocks,
with an encoding dimension of 256, and 4 attention heads. The seq2seq model,
like CTC, requires the word embeddings to compute the word-level log
probabilities. However, unlike CTC, the word embeddings are also input to the
decoder, $g(\cdot)$.

The attention mechanism in the seq2seq model implicitly learns an alignment,
thus the final optimization objective is simply the log probability of the word
sequence:
\begin{equation}
  \label{eq-s2s}
    \log P(\mathbf{Y}|\mathbf{X}, \mathbf{W}) = \sum_{n=0}^N \log P(y_n | y_{<n}, \mathbf{X},\, \mathbf{W}).
\end{equation}
where $y_0$ is a special token indicating the beginning of the transcription.

\begin{table*}[ht!]
    \centering
    \caption{\label{tab-fulllex}Comparison of our word-level approach against
    a word piece baseline on the full 960 hour dataset, with and without
    language model (LM) decoding. All models have an overall stride of 8 except
    rows with $s=\!16$, which have a stride of 16.}
    \vspace{2mm}
    \begin{tabular}{l c c c S[table-format=2.2, round-mode=places, round-precision=1] S[table-format=2.2, round-mode=places, round-precision=1] S[table-format=2.2, round-mode=places, round-precision=1] S[table-format=2.2, round-mode=places, round-precision=1]}
        \toprule
        \multirow{2}{*}{Model} & \multirow{2}{*}{LM} & $|\mathcal{D}|$ & $|\mathcal{D}|$ & \multicolumn{2}{c}{Dev WER} & \multicolumn{2}{c}{Test WER}  \\
                               &                     & train & test & \multicolumn{1}{c}{clean} & \multicolumn{1}{c}{other} & \multicolumn{1}{c}{clean} & \multicolumn{1}{c}{other} \\
        \midrule
        word piece Trans. CTC~\cite{synnaeve2019e2e} & None & - & - & 2.98 & 7.36 & 3.18 & 7.49 \\
        word piece Trans. CTC~\cite{synnaeve2019e2e} & 4-gram & - & - & 2.51 & 6.21 & 2.92 & 6.65 \\
        word piece Trans. seq2seq~\cite{synnaeve2019e2e} & None & - & - & 2.56 & 6.65 & 3.05 & 7.01 \\
        word piece Trans. seq2seq~\cite{synnaeve2019e2e} & 6-gram & - & - & 2.28 & 5.88 & 2.58 & 6.15 \\
       \midrule
       \midrule
        word-level Trans.  CTC              & None   & 89k & 89k  & 2.91 & 7.49 & 3.15 & 7.45 \\
        word-level Trans.  CTC              & 4-gram & 89k & 89k  & 2.69 & 6.62 & 2.88 & 6.73 \\
        word-level Trans.  CTC              & 4-gram & 89k & 200k & 2.59 & 6.60 & 2.91 & 6.72 \\
        word-level Trans.  seq2seq              & None   & 89k & 89k  & 2.66 & 6.48 & 2.87 & 6.69 \\
        word-level Trans.  seq2seq              & 4-gram & 89k & 89k  & 2.58 & 5.96 & 2.78 & 6.18 \\
        word-level Trans.  seq2seq              & 4-gram & 89k & 200k & 2.53 & 5.96 & 3.02 & 6.28 \\
        word-level Trans.  CTC, $s\!=\!16$  & None   & 89k & 89k  & 2.97 & 7.69 & 3.24 & 7.73 \\
        word-level Trans.  CTC, $s\!=\!16$  & 4-gram & 89k & 89k  & 2.66 & 6.75 & 2.98 & 6.97 \\
        word-level Trans.  CTC, $s\!=\!16$  & 4-gram & 89k & 200k & 2.55 & 6.72 & 2.98 & 6.96 \\
        word-level Trans.  seq2seq, $s\!=\!16$  & None   & 89k & 89k  & 2.81 & 7.14 & 3.29 & 7.48 \\
        word-level Trans.  seq2seq, $s\!=\!16$  & 4-gram & 89k & 89k  & 2.65 & 6.39 & 2.93 & 6.90 \\
        word-level Trans.  seq2seq, $s\!=\!16$  & 4-gram & 89k & 200k & 2.58 & 6.35 & 2.99 & 6.86 \\
       \bottomrule
    \end{tabular}
\end{table*}

\subsection{Scalable Training via Word Sampling}
\label{sec:sampling}

The resources (computation and memory) required to estimate the posteriors
in~\voireq{eq-wordscore} scale linearly with the number of words in the lexicon
${\cal D}$. As is, word-level approaches do not scale well to lexicons with
more than a few thousand. To circumvent this scaling issue, we use a simple
sampling approach: for each sample $(\mathbf{X}, \mathbf{Y})$ (or batch of
samples) we consider a dictionary $\tilde{\cal D}$ with the size $|\tilde{\cal
D}| \ll |{\cal D}|$ fixed beforehand. Let $\mathcal{D}_\mathbf{Y}$ be the set
of unique labels in $\mathbf{Y}$. We construct $\tilde{\cal D}$ by combining
$\mathcal{D}_\mathbf{Y}$ and a set of labels uniformly sampled from ${\cal D}
\backslash \cal{D}_{\mathbf{Y}}$. All operations described in previous sections
then remain the same, but performed using $\tilde{\cal D}$. For simplicity, in
practice we construct $\mathcal{D}_\mathbf{Y}$ from a mini-batch of samples
rather than just a single example. We will show in \voirsec{sec-experiments}
that this sampling approach is effective in practice.

\subsection{Inference and Language Model Decoding}

For fast inference, the embeddings $\mathbf{W}$ can be computed in advance
only \emph{once}, if the inference dictionary is known beforehand, and
fixed. Note that the inference dictionary may or may not be the same as
the training dictionary.

As our approach is word-based, there is no need for a decoding procedure at
inference time -- compared to sub-word unit-based approaches -- to obtain a
sequence of meaningful words. However, leveraging a word language model trained
on a large text corpus can still help to further decrease the word error rate
(see \voirsec{sec-experiments}). We implemented a beam-search decoder which
optimizes the following objective:
\begin{equation}
  \label{eq-beam-search}
  \log P(\mathbf{Y}|\mathbf{X}, \mathbf{W}) + \alpha\, \log P_{LM}(\mathbf{Y}) + \beta |\mathbf{Y}|\,,
\end{equation}
where $\log P_{LM}(\cdot)$ is the log-likelihood of the language model,
$\alpha$ is the weight of the language model, and $\beta$ is a word insertion
weight. Given the acoustic and language models operate at the same granularity
(words), the decoder is much simpler to implement than a traditional speech
recognition decoder. The decoder is an iterative beam search procedure which
tracks hypotheses with the highest probability in the following steps:
\begin{itemize}
\item Each hypothesis is augmented with a word in ${\cal D}$. For efficiency,
    only the top $K$ words, according to the acoustic model likelihood, are
    considered.
\item The individual hypothesis scores are updated according to the acoustic
    and language model scores, as well as the word insertion score.
\item The top $B$ hypotheses are retained and the remaining hypotheses are
    discarded.
\end{itemize}
Both $K$ and $B$ are hyper-parameters which trade-off accuracy and efficiency
and can be set by the user.

\section{Experiments}
\label{sec-experiments}

We perform experiments on the LibriSpeech corpus -- 960 hours of speech
collected from open domain audio books~\cite{panayotov2015librispeech}. We
consider two settings. In the first, models are trained on all of the available
training data. In the second, we limit the training data to a cleaner 100 hour
subset (``train-clean-100'') in order to study our model in a lower resource
setting. All hyper-parameters are tuned according to the word error rates on
the standard validation sets. Final test set performance is reported for both
the \textsc{clean} and \textsc{other} settings (the latter being a subset of
the data with noisier utterances). We use log-mel filterbanks as features to
the acoustic model, with 80 filters of size 25ms, stepped by 10ms.  No speaker
adaptation was performed. We use SpecAugment for data augmentation, following
the recipe in~\citet{park2019specaugment}.

Unless otherwise stated, the lexicon ${\cal D}$ at training time contains all
the words in the training set (around 89k words) and we use a 5k word
lexicon at each iteration sampled as described in Section~\ref{sec:sampling}.
Characters used to build the word embeddings include all English letters (a-z)
and the apostrophe, augmented by special tokens \textsc{<pad>},
\textsc{<blank>} (for CTC) and \textsc{<eos>} (for seq2seq), as described in
\voirsec{sec-wordembed}. When decoding with a language model, we use the
standard LibriSpeech 4-gram LM, which contains 200k unigrams.  For some of the
word piece models, we decode with a 6-gram word piece LM designed to match the
perplexity of the 4-gram word level model. The word piece 6-gram is trained on
the same LibriSpeech training text as the 4-gram.  We use the open source
\textsc{wav2letter++} toolkit~\cite{pratap2018} to perform our experiments. We
train with Stochastic Gradient Descent, with a mini-batch size of $128$ samples
split evenly across eight GPUs. We use the Transformer encoder as described in
\voirsec{sec-transformer} for both CTC and seq2seq models and the Transformer
decoder for seq2seq models as described in \voirsec{sec-s2s}.

Our initial runs with the CTC word-level approach were unsuccessful -- the loss
and WER diverged after a few iterations. We found that the norm of the
output of the acoustic and word embedding models were slowly growing in
magnitude causing numerical instability and ultimately leading to divergence.
We circumvent the issue by constraining the norm of the embeddings to lie in an
$L_2$-ball. Keeping $\|f_t^{am}\|_2 \le 5$ and $\|f_t^{wd}\|_2 \le 5$ for both
the acoustic and word model embeddings stabilized training. All of our
subsequent experiments were performed with this constraint.

In~\voirtbl{tab-fulllex}, we compare our word-level approach with word pieces,
which are commonly used with seq2seq models in speech recognition and machine
translation. On the 960 hour training set, the word piece models contain 10k
tokens computed from the \textit{SentencePiece}
toolkit~\cite{kudo2018sentencepiece}. We also show
in~\voirtbl{tab-fulllex} that a model with a stride of 16 gives comparable WERs
to a model with a stride of 8 while being faster to train and decode, and
having a smaller memory footprint. We experimented with word piece models and
higher strides and found that the results degraded much more beyond a stride of
8.

We also compare the word-level approach to word piece models and other prior work
on the 100 hour subset. The lexicon-free word piece baseline uses a word piece
set of 5k tokens learned from the transcriptions of ``train-clean-100''. We
cross validated over different size word piece sets and found 5k to be optimal.
The word-level models are trained with a lexicon containing 89k words and
evaluated with a lexicon containing 200k words. The 200k lexicon is the lexicon
from the default LibriSpeech language model.

In Table~\ref{tab:train100h}, we show that our end-to-end models, both
word-level and word piece, substantially outperform prior work on the 100 hour
subset ``train-clean-100''. We also see that the word-level models consistently
outperform the word piece models both with and without language model decoding.
This suggests the word-level approach has an advantage over word pieces in the
lower resource setting. We also observe that the word-level model more
substantially improves over word piece models without the use of an external
LM. This may be due to several factors including the benefit of having a
lexicon built into the model, and that word-level models can learn a stronger
implicit LM.

\begin{table}
    \setlength{\tabcolsep}{2pt}
    \small
    \centering
    \caption{\label{tab:train100h}Comparison of our word-level approaches against
    a word piece baseline and prior work on the full 100 hour dataset, with and without
    language model (LM) decoding.}
    \vspace{2mm}
    \begin{tabular}{l c c c c c}
        \toprule
        \multirow{2}{*}{Model} & \multirow{2}{*}{LM} & \multicolumn{2}{c}{Dev WER} & \multicolumn{2}{c}{Test WER}  \\
                               &                     & clean & other & clean & other \\
       \midrule
        \citet{luscher2019rwth} hybrid  & 4-gram  & 5.0 & 19.5  & 5.8  & 18.6 \\
        \citet{luscher2019rwth} seq2seq & None & 14.7 & 38.5  & 14.7  & 40.8 \\
        \citet{irie2019model} seq2seq   & None & 12.7 & 33.9  & 12.9  & 35.5 \\
       \midrule
        word piece seq2seq     & None   & 9.0  & 22.8 & 9.5  & 23.3 \\
        word piece seq2seq     & 6-gram & 8.3  & 21.2 & 9.2  & 22.0 \\
        word piece CTC         & None   & 12.4 & 27.7 & 12.8 & 28.7 \\
        word piece CTC         & 6-gram & 9.7  & 22.9 & 10.3 & 24.0 \\
       \midrule
       \midrule
        word-level seq2seq  & None   & 7.2 & 21.2 & 8.6 & 21.9 \\
        word-level seq2seq  & 4-gram & 7.3 & 19.5 & 8.0 & 20.4 \\
        word-level CTC      & None   & 8.0 & 21.0 & 7.7 & 21.4 \\
        word-level CTC      & 4-gram & \textbf{6.3} & \textbf{19.1} & \textbf{6.8} & \textbf{19.4} \\
       \bottomrule
    \end{tabular}
\end{table}

\begin{figure*}
  \centering
    \subfloat[Dev \textsc{clean}]{\label{fig:sampling-clean}\includegraphics[width=0.4\linewidth]{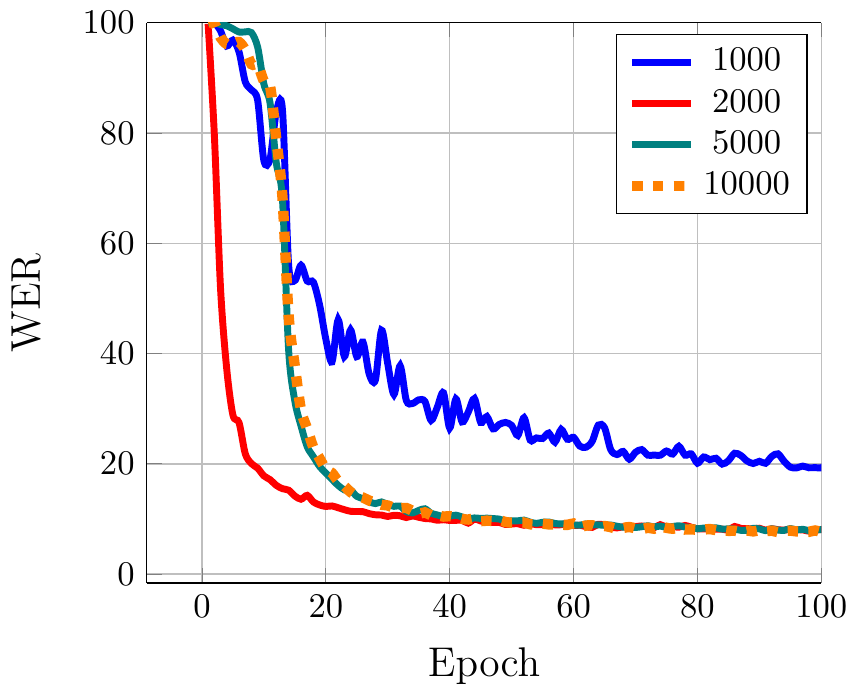}}
    \hspace{5mm}
    \subfloat[Dev \textsc{other}]{\label{fig:sampling-other}\includegraphics[width=0.4\linewidth]{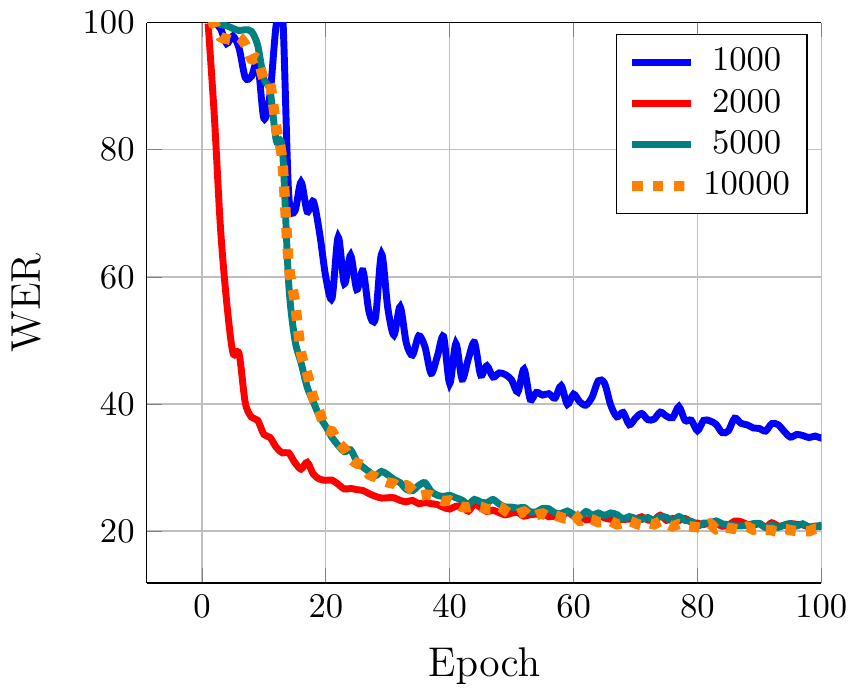}}
  \caption{
    \label{fig:sampling}
    Validation WER with respect to number of training epochs, on the (a)
    \textsc{clean} and (b) \textsc{other} conditions. Training was with
    CTC. We show the effect of the number of sampled words (1000, 2000,
    5000 and 10000) used in the CTC normalization (see~\voireq{eq-wordscore}).}
\end{figure*}

\subsection{Effect of Sampling}

We compare the validation WER on both the \textsc{clean} and \textsc{other}
conditions, for a different number of sampled words from the lexicon ${\cal
D}$, as explained in \voirsec{sec:sampling}. We report results for the CTC
trained model using the Time-Depth Separable (TDS) convolution architecture
of~\citet{hannun2019sequence} with 9 TDS blocks -- the acoustic model used in
most of our preliminary experiments.

As we see in \voirfig{fig:sampling}, too few sampled examples makes the training
unstable, and slow to converge. We do not see any advantage in having more
than 5k sampled words per batch. We observe in general that the number of
sampled examples may affect the beginning of training -- but in our experience,
when the sample set is large enough ($\ge$2000), all experiments converge to a
similar word error rate.

\subsection{Out-of-Vocabulary Words}

\begin{table*}
    \centering
    \caption{\label{tab-oov-neighbors} Word neighbors for selected OOVs using
      letter edit distance and Euclidean distance in the word model
      embedding space. The word model was trained with CTC on LibriSpeech
      100h. The training vocabulary contains 39k words, and neighbors are
      computed with the full LM dictionary (200k words). Words in
      \emph{italic} are OOV (not seen at training time).}  \small
    \vspace{2mm}
    \begin{tabular}{cc cc cc cc cc}
        \toprule
        \multicolumn{2}{c}{\emph{amulet}} & \multicolumn{2}{c}{\emph{cinderella's}} & \multicolumn{2}{c}{\emph{mittens}} & \multicolumn{2}{c}{\emph{snobbish}} & \multicolumn{2}{c}{\emph{welsh}} \\
(Edit)        & (Embed.) & (Edit)        & (Embed.) & (Edit)        & (Embed.) & (Edit)        & (Embed.) & (Edit)        & (Embed.) \\
\midrule
\emph{mulet} & emulate &         \emph{cinderellas} & \emph{cinderellas} &    \emph{kittens} & \emph{mattins} & \emph{snobbism} & \emph{snobbism} &     \emph{weesh} & \emph{walsh}      \\
\emph{amulets} & \emph{amelot} & cinderella & cinderella &                    \emph{mitten} & \emph{mittin} &   \emph{snobbishly} & \emph{snobby} &     \emph{walsh} & \emph{welch}      \\
\emph{amile} & \emph{omelet} &   \emph{cantrell's} & \emph{kinsella's} &      \emph{tittens} & \emph{miltons} & \emph{bobbish} & \emph{snobbs} &        \emph{wesh} & \emph{walch}       \\
\emph{amplest} & \emph{emilet} & \emph{banderillas} & \emph{chaudoreille's} & \emph{ritten} & \emph{middens} &  \emph{snobbs} & \emph{snobbe} &         \emph{welshy} & \emph{welsher}   \\
\emph{ambles} & \emph{amoret} &  \emph{calderwell's} & \emph{ghirlandajo's} & \emph{mittin} & \emph{mitten} &   \emph{snobbery} & \emph{snobbishness} & \emph{welse} & \emph{wesh}       \\
\emph{mulct} & \emph{amalek} &   \emph{kinsella's} & \emph{sganarelle's} &    \emph{battens} & \emph{mitton} &  \emph{nobis} & \emph{snubby} &          \emph{welch} & \emph{walshe}     \\
\emph{amoret} & \emph{amulets} & \emph{cinereous} & \emph{sinclair's} &       \emph{fattens} & \emph{muttons} & \emph{nobbs} & \emph{snob's} &          \emph{weals} & \emph{walth}      \\
\emph{roulet} & armlet &         \emph{cordelia's} & \emph{cibolero's} &      pattens & \emph{matins} &         \emph{sourish} & \emph{snob} &          \emph{selah} & \emph{welshy}     \\
       \bottomrule
    \end{tabular}
\end{table*}

In~\voirtbl{tab-fulllex}, we show that training with the complete training
lexicon (89k words), and then decoding with the language model lexicon shipped
with LibriSpeech (200k words) yields similar WERs as decoding with the 89k
lexicon. At evaluation time, we simply extended the dictionary by computing the
unseen word embeddings with the letter-based word model.
We consider as out-of-vocabulary (OOV) words which are considered at
inference, but not present at training time.

To further study the extent to which new words can be added to the lexicon, we
trained word piece and word-level models on the ``train-clean-100'' subset of
LibriSpeech. The lexicon for the word-level models was chosen to contain only
the words which occur in the transcripts of this subset, of which there are
about 34k. Given the word model receives letters as input, we
compare the neighbors in the word model output embedding space to neighbors
found using letter edit distance. \voirtbl{tab-oov-neighbors} shows a
comparison of the eight nearest neighbors for select OOV words using these two
metrics. The neighbors found by the word model overlap with those found
using letter edit distance. However, using only edit distance results in words
which are spelled similarly but with clearly differing pronunciations as close
neighbors. The word model is able to correctly discard these from the set of
close neighbors. We also find that some morphology is learned by the word
model (e.g. with ``'s'' in ``cinderella's'').

\begin{figure}
  \centering
    \subfloat[Dev \textsc{clean}]{\label{fig:oov-wer-clean}\includegraphics[width=0.4\linewidth]{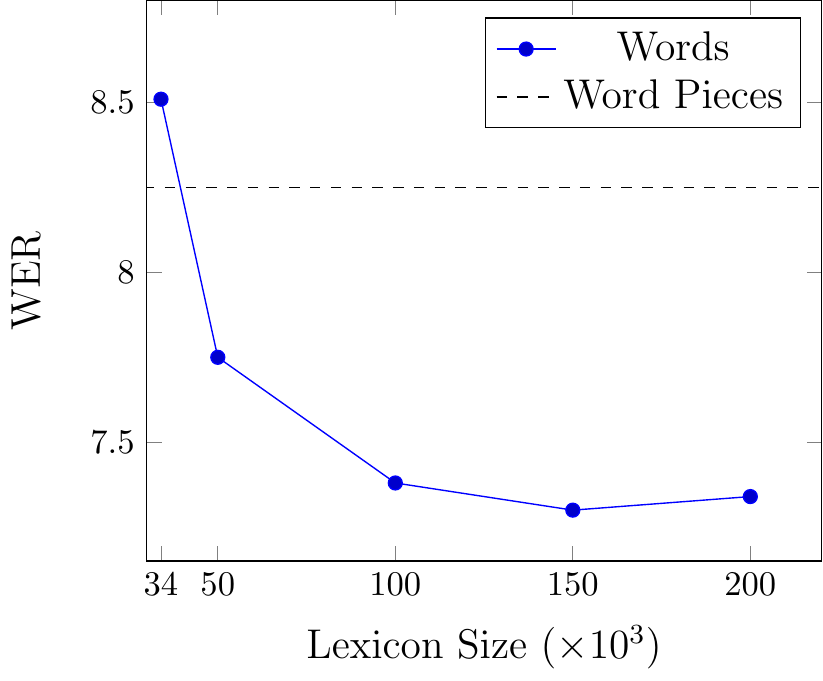}}
    \subfloat[Dev \textsc{other}]{\label{fig:oov-wer-other}\includegraphics[width=0.4\linewidth]{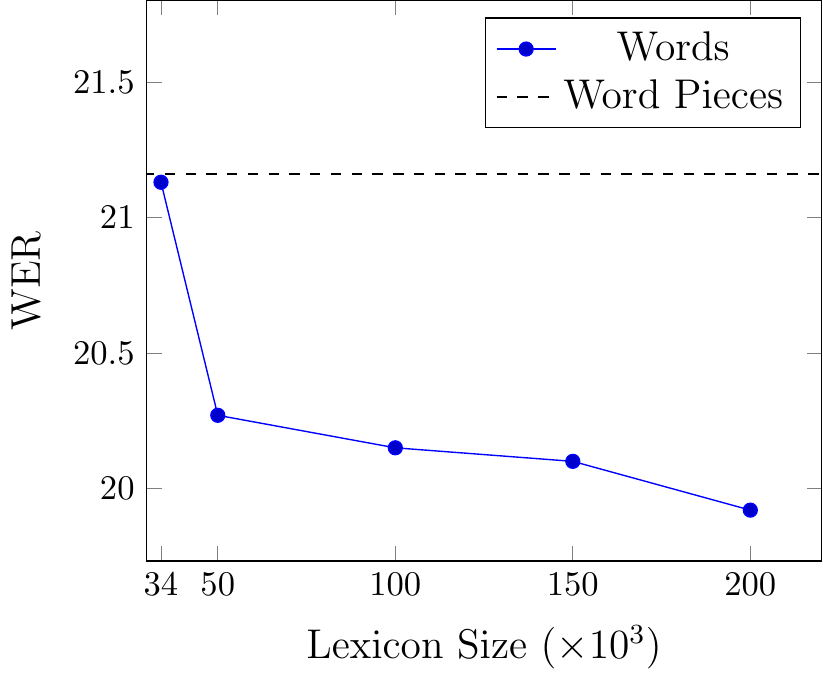}}
  \caption{
    \label{fig:oov-wer}
    The WER on the (a) \textsc{clean} and (b) \textsc{other} development sets
    as a function of the lexicon size used when evaluating the word-level
    model. We use a 4-gram word LM for the word model and a 6-gram word piece
    LM for the word piece model. The word piece model uses 5k tokens but is
    lexicon-free.
}
\end{figure}

In Figure~\ref{fig:oov-wer} we examine the word error rate as we increase
the size of the lexicon used when evaluating the development set. When
using the same 34k lexicon used at training time for evaluation, the
word-level and word piece models perform comparably. However, as we
increase the lexicon size the word-level model improves and consistently
outperforms the word piece model. The larger lexicons are chosen by keeping
the top words by frequency computed from the LibriSpeech language model
training data.

\begin{figure*}[ht!]
    \centering
  \subfloat[OOV Precision]{\label{fig:oov-precision}\includegraphics[width=0.4\linewidth]{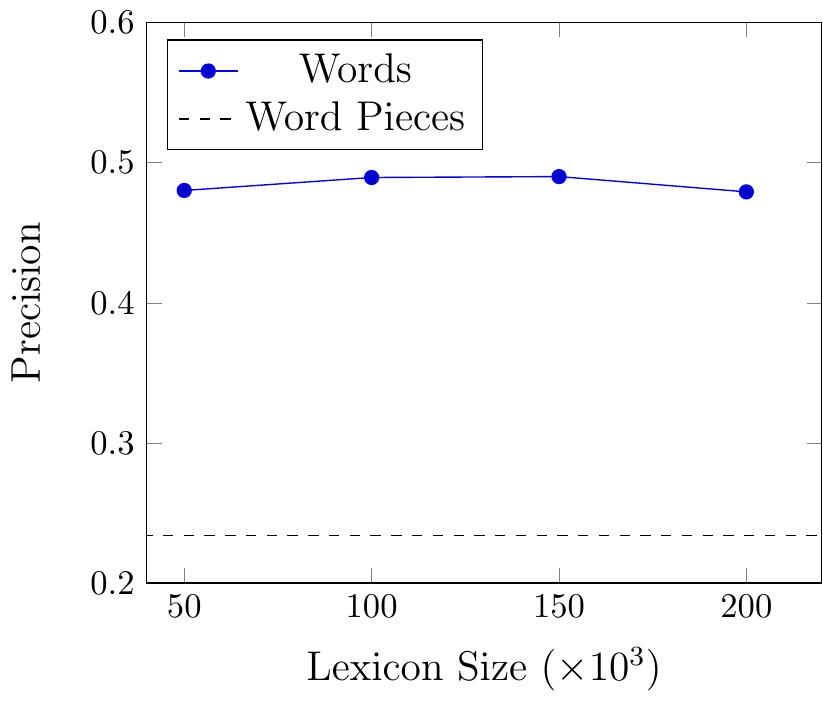}}
    \hspace{5mm}
  \subfloat[OOV Recall]{\label{fig:oov-recall}\includegraphics[width=0.4\linewidth]{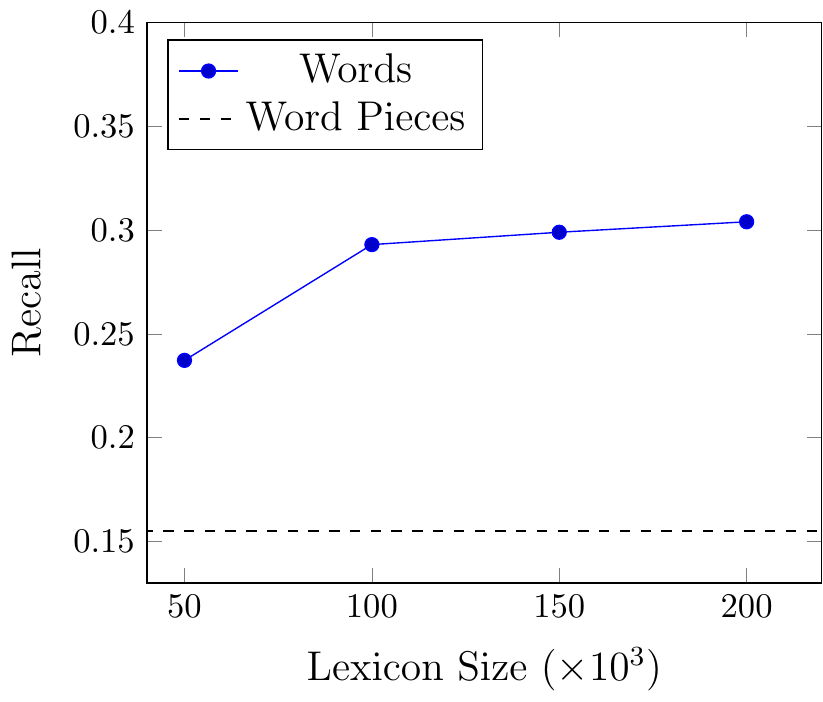}}
  \caption{
    \label{fig:oov}
    Out-of-vocabulary (a) precision and (b) recall as a function of the lexicon
    size used when evaluating the word-level model. An OOV is any token which does
    not appear in the set of 34k unique tokens present in the 100 hour training
    set. The word piece model uses 5k tokens but is lexicon-free.
}
\end{figure*}

We also study the performance of the word-level and word piece models limited
to out-of-vocabulary words. To do so, we first find the minimum word edit
distance alignment of the models best prediction to the reference transcript.
Using this alignment, we define the OOV recall as the number of correctly
classified OOV words over the total number of OOVs in the reference
transcripts. Similarly, we define the OOV precision as the number of correctly
classified OOV words over the total number of predicted OOVs.

For example, given the reference \texttt{the cat sat} and the prediction
\texttt{cat sat} we compute the following alignment (where \texttt{the}
corresponds to a deletion):
\begin{align*}
    \textrm{Reference:}\quad    &\texttt{the} \quad \texttt{cat} \quad \texttt{sat} \\
    \textrm{Prediction:}\quad   &\texttt{***}  \quad \texttt{cat} \quad \texttt{sat}
\end{align*}
If both \texttt{the} and \texttt{sat} are OOV words, then the precision is
$1$ and the recall is $1/2$, for this example.

For both word-level and word piece models we define an OOV as a word which does
not appear in any of the training set transcriptions (i.e.  any word not in the
34k lexicon). We compute OOV precision and recall on the combined
\textsc{clean} and \textsc{other} development sets.

As seen in Figure~\ref{fig:oov}, the word model achieves better OOV precision
(a) and recall (b) than the corresponding word piece model. We also see that
the word model improves in recall as we increase the lexicon size without a
loss in precision. This may explain the improved word error rate of the word
model with larger lexicon sizes as seen in Figure~\ref{fig:oov-wer}.

\subsection{Efficiency}
We compared the wall-clock time of word-based acoustic models.
For decoding, the decoding time per utterance is 0.539 seconds at stride 8
(80ms), and 0.323 seconds at stride 16 (160ms). Just the forward of the
acoustic model for a batch of 5 utterances takes 85ms and 60ms for stride 8 and
stride 16 respectively. The whole batch (forward, criterion, backward,
optimizer) takes 469ms (stride 8) vs. 426ms (stride 16), so the gain is mostly
at inference time (forward). Also note that, in practice, we can fit larger
batches in GPU memory with stride 16 than 8, and all in all (even with the
additional FLOPS) the throughput will be higher for stride 16, but this was ran
at the same batch size for comparison.

Compared to word piece models, the softmax of the word-level model is more
computationally expensive at inference time. However, with a vocabulary of 89k and stride 8
model, computing the last layer represents only $11\%$ of the overall
forward time (to compare with $9\%$ with a vocabulary of 10k -- as in our
word piece models -- and $15\%$ with 200k). Word-level models have the
advantage of allowing stride 16 with little loss in WER performance (see
\voirtbl{tab-fulllex}), and twice faster decoding time.

\section{Conclusion}

Direct-to-word speech recognition poses an exciting opportunity to simplify ASR
models as well as make them more accurate and more efficient. However, this
line of research has received little attention due to the difficulty of the
problem. Either the model requires massive training sets to learn reasonably
sized lexicons or the lexicon is small and unchangeable.

We have demonstrated that a direct-to-word approach for speech recognition is
not only possible but promising. Our model gains several advantages from
predicting words directly including (1)~an improvement in WER over word piece
baselines (2)~a model which can operate at a larger stride (lower frame-rate)
and hence is more efficient and (3)~a simpler beam search decoder which
integrates easily with an externally trained language model. Key to our
approach is a letter to word embedding model which can be jointly trained with
an acoustic embedding model.  To make this efficient, especially for large
vocabulary sizes, we used a simple sampling-based mechanism to compute the
normalization term needed when training.

We have shown that our approach can be used seamlessly with two commonly used
end-to-end models for speech recognition -- a CTC trained model and an
encoder-decoder model with attention. We validated our models on LibriSpeech, a
standard benchmark in speech recognition. On this data set, the
direct-to-word model achieves competitive word error rates. We also showed that
the direct-to-word model has a stronger advantage over a well tuned word piece
baseline in a lower resource setting. In fact, in this setting our model
achieves a state-of-the-art result for end-to-end models. Finally, we demonstrated
that our model can accurately predict words never seen in the training set
transcriptions and gave evidence that the word-level model generalizes to
out-of-vocabulary words better than a word piece model.

\begin{small}
	\bibliography{refs}

\begin{thebibliography}{42}
\providecommand{\natexlab}[1]{#1}
\providecommand{\url}[1]{\texttt{#1}}
\expandafter\ifx\csname urlstyle\endcsname\relax
  \providecommand{\doi}[1]{doi: #1}\else
  \providecommand{\doi}{doi: \begingroup \urlstyle{rm}\Url}\fi

\bibitem[Amodei et~al.(2016)Amodei, Ananthanarayanan, Anubhai, Bai, Battenberg,
  Case, Casper, Catanzaro, Cheng, Chen, et~al.]{amodei2016}
Dario Amodei, Sundaram Ananthanarayanan, Rishita Anubhai, Jingliang Bai, Eric
  Battenberg, Carl Case, Jared Casper, Bryan Catanzaro, Qiang Cheng, Guoliang
  Chen, et~al.
\newblock Deep speech 2: End-to-end speech recognition in english and mandarin.
\newblock In \emph{International Conference on Machine Learning (ICML)}, pages
  173--182, 2016.

\bibitem[Audhkhasi et~al.(2017)Audhkhasi, Ramabhadran, Saon, Picheny, and
  Nahamoo]{audhkhasi2017direct}
Kartik Audhkhasi, Bhuvana Ramabhadran, George Saon, Michael Picheny, and David
  Nahamoo.
\newblock Direct acoustics-to-word models for english conversational speech
  recognition.
\newblock \emph{arXiv preprint arXiv:1703.07754}, 2017.

\bibitem[Ba et~al.(2016)Ba, Kiros, and Hinton]{ba2016layer}
Jimmy~Lei Ba, Jamie~Ryan Kiros, and Geoffrey~E Hinton.
\newblock Layer normalization.
\newblock \emph{arXiv preprint arXiv:1607.06450}, 2016.

\bibitem[Bahdanau et~al.(2014)Bahdanau, Cho, and Bengio]{bahdanau2014neural}
Dzmitry Bahdanau, Kyunghyun Cho, and Yoshua Bengio.
\newblock Neural machine translation by jointly learning to align and
  translate.
\newblock In \emph{International Conference on Learning Representations
  (ICLR)}, 2014.

\bibitem[Bahdanau et~al.(2016)Bahdanau, Chorowski, Serdyuk, Brakel, and
  Bengio]{bahdanau2016icassp}
Dzmitry Bahdanau, Jan Chorowski, Dmitriy Serdyuk, Philemon Brakel, and Yoshua
  Bengio.
\newblock End-to-end attention-based large vocabulary speech recognition.
\newblock In \emph{International Conference on Acoustics, Speech and Signal
  Processing (ICASSP)}, pages 4945--4949. IEEE, 2016.

\bibitem[Bengio and Heigold(2014)]{bengio2014word}
Samy Bengio and Georg Heigold.
\newblock Word embeddings for speech recognition.
\newblock In \emph{Fifteenth Annual Conference of the International Speech
  Communication Association}, 2014.

\bibitem[Bisani and Ney(2008)]{bisani2008joint}
Maximilian Bisani and Hermann Ney.
\newblock Joint-sequence models for grapheme-to-phoneme conversion.
\newblock \emph{Speech communication}, 50\penalty0 (5):\penalty0 434--451,
  2008.

\bibitem[Bojanowski et~al.(2017)Bojanowski, Grave, Joulin, and
  Mikolov]{bojanowski2017enriching}
Piotr Bojanowski, Edouard Grave, Armand Joulin, and Tomas Mikolov.
\newblock Enriching word vectors with subword information.
\newblock \emph{Transactions of the Association for Computational Linguistics},
  5:\penalty0 135--146, 2017.

\bibitem[Chan et~al.(2016)Chan, Jaitly, Le, and Vinyals]{chan2016listen}
William Chan, Navdeep Jaitly, Quoc Le, and Oriol Vinyals.
\newblock Listen, attend and spell: A neural network for large vocabulary
  conversational speech recognition.
\newblock In \emph{Acoustics, Speech and Signal Processing (ICASSP), 2016 IEEE
  International Conference on}, pages 4960--4964. IEEE, 2016.

\bibitem[Cho et~al.(2014)Cho, Van~Merri{\"e}nboer, Gulcehre, Bahdanau,
  Bougares, Schwenk, and Bengio]{cho2014learning}
Kyunghyun Cho, Bart Van~Merri{\"e}nboer, Caglar Gulcehre, Dzmitry Bahdanau,
  Fethi Bougares, Holger Schwenk, and Yoshua Bengio.
\newblock Learning phrase representations using rnn encoder-decoder for
  statistical machine translation.
\newblock 2014.

\bibitem[Collobert et~al.(2016)Collobert, Puhrsch, and Synnaeve]{collobert2016}
Ronan Collobert, Christian Puhrsch, and Gabriel Synnaeve.
\newblock Wav2letter: an end-to-end convnet-based speech recognition system.
\newblock \emph{arXiv preprint arXiv:1609.03193}, 2016.

\bibitem[Collobert et~al.(2019)Collobert, Hannun, and Synnaeve]{collobert2019}
Ronan Collobert, Awni Hannun, and Gabriel Synnaeve.
\newblock A fully differentiable beam search decoder.
\newblock In \emph{International Conference on Machine Learning {(ICML)}},
  2019.

\bibitem[Dauphin et~al.(2017)Dauphin, Fan, Auli, and Grangier]{dauphin2017}
Yann~N Dauphin, Angela Fan, Michael Auli, and David Grangier.
\newblock Language modeling with gated convolutional networks.
\newblock In \emph{International Conference on Machine Learning {(ICML)}},
  pages 933--941, 2017.

\bibitem[Fan et~al.(2019)Fan, Grave, and Joulin]{fan2019reducing}
Angela Fan, Edouard Grave, and Armand Joulin.
\newblock Reducing transformer depth on demand with structured dropout.
\newblock \emph{arXiv preprint arXiv:1909.11556}, 2019.

\bibitem[Gibson and Hain(2006)]{gibson2006hypothesis}
Matthew Gibson and Thomas Hain.
\newblock Hypothesis spaces for minimum bayes risk training in large vocabulary
  speech recognition.
\newblock In \emph{Ninth international conference on spoken language
  processing}, 2006.

\bibitem[Graves and Jaitly(2014)]{graves2014towards}
Alex Graves and Navdeep Jaitly.
\newblock Towards end-to-end speech recognition with recurrent neural networks.
\newblock In \emph{International Conference on Machine Learning {(ICML)}},
  pages 1764--1772, 2014.

\bibitem[Graves et~al.(2006)Graves, Fern{\'a}ndez, Gomez, and
  Schmidhuber]{graves2006}
Alex Graves, Santiago Fern{\'a}ndez, Faustino Gomez, and J{\"u}rgen
  Schmidhuber.
\newblock Connectionist temporal classification: labelling unsegmented sequence
  data with recurrent neural networks.
\newblock In \emph{International Conference on Machine Learning {(ICML)}},
  pages 369--376, 2006.

\bibitem[Hannun et~al.(2019)Hannun, Lee, Xu, and Collobert]{hannun2019sequence}
Awni Hannun, Ann Lee, Qiantong Xu, and Ronan Collobert.
\newblock Sequence-to-sequence speech recognition with time-depth separable
  convolutions.
\newblock \emph{arXiv preprint arXiv:1904.02619}, 2019.

\bibitem[Irie et~al.(2019)Irie, Prabhavalkar, Kannan, Bruguier, Rybach, and
  Nguyen]{irie2019model}
Kazuki Irie, Rohit Prabhavalkar, Anjuli Kannan, Antoine Bruguier, David Rybach,
  and Patrick Nguyen.
\newblock On the choice of modeling unit for sequence-to-sequence speech
  recognition.
\newblock In \emph{Interspeech}, 2019.

\bibitem[Kanthak and Ney(2002)]{kanthak2002context}
Stephan Kanthak and Hermann Ney.
\newblock Context-dependent acoustic modeling using graphemes for large
  vocabulary speech recognition.
\newblock In \emph{2002 IEEE International Conference on Acoustics, Speech, and
  Signal Processing}, volume~1, pages I--845. IEEE, 2002.

\bibitem[Killer et~al.(2003)Killer, Stuker, and Schultz]{killer2003grapheme}
Mirjam Killer, Sebastian Stuker, and Tanja Schultz.
\newblock Grapheme based speech recognition.
\newblock In \emph{Eighth European Conference on Speech Communication and
  Technology}, 2003.

\bibitem[Kim et~al.(2016)Kim, Jernite, Sontag, and Rush]{kim2016character}
Yoon Kim, Yacine Jernite, David Sontag, and Alexander~M Rush.
\newblock Character-aware neural language models.
\newblock In \emph{Thirtieth AAAI conference on artificial intelligence}, 2016.

\bibitem[Kudo and Richardson(2018)]{kudo2018sentencepiece}
Taku Kudo and John Richardson.
\newblock {S}entence{P}iece: A simple and language independent subword
  tokenizer and detokenizer for neural text processing.
\newblock \emph{arXiv preprint arXiv:1808.06226}, 2018.

\bibitem[Labeau and Allauzen(2017)]{labeau2017character}
Matthieu Labeau and Alexandre Allauzen.
\newblock Character and subword-based word representation for neural language
  modeling prediction.
\newblock In \emph{Proceedings of the First Workshop on Subword and Character
  Level Models in NLP}, pages 1--13, 2017.

\bibitem[Li et~al.(2018)Li, Ye, Das, Zhao, and Gong]{li2018advancing}
Jinyu Li, Guoli Ye, Amit Das, Rui Zhao, and Yifan Gong.
\newblock Advancing acoustic-to-word ctc model.
\newblock In \emph{2018 IEEE International Conference on Acoustics, Speech and
  Signal Processing (ICASSP)}, pages 5794--5798. IEEE, 2018.

\bibitem[Likhomanenko et~al.(2019)Likhomanenko, Synnaeve, and
  Collobert]{likhomanenko2019}
Tatiana Likhomanenko, Gabriel Synnaeve, and Ronan Collobert.
\newblock Who needs words? lexicon-free speech recognition.
\newblock \emph{arXiv preprint arXiv:1904.04479}, 2019.

\bibitem[Ling et~al.(2015{\natexlab{a}})Ling, Lu{\'\i}s, Marujo, Astudillo,
  Amir, Dyer, Black, and Trancoso]{ling2015finding}
Wang Ling, Tiago Lu{\'\i}s, Lu{\'\i}s Marujo, Ram{\'o}n~Fernandez Astudillo,
  Silvio Amir, Chris Dyer, Alan~W Black, and Isabel Trancoso.
\newblock Finding function in form: Compositional character models for open
  vocabulary word representation.
\newblock \emph{arXiv preprint arXiv:1508.02096}, 2015{\natexlab{a}}.

\bibitem[Ling et~al.(2015{\natexlab{b}})Ling, Trancoso, Dyer, and
  Black]{ling2015character}
Wang Ling, Isabel Trancoso, Chris Dyer, and Alan~W Black.
\newblock Character-based neural machine translation.
\newblock \emph{arXiv preprint arXiv:1511.04586}, 2015{\natexlab{b}}.

\bibitem[L{\"u}scher et~al.(2019)L{\"u}scher, Beck, Irie, Kitza, Michel, Zeyer,
  Schl{\"u}ter, and Ney]{luscher2019rwth}
Christoph L{\"u}scher, Eugen Beck, Kazuki Irie, Markus Kitza, Wilfried Michel,
  Albert Zeyer, Ralf Schl{\"u}ter, and Hermann Ney.
\newblock Rwth asr systems for librispeech: Hybrid vs attention-w/o data
  augmentation.
\newblock In \emph{Interspeech}, 2019.

\bibitem[Maas et~al.(2015)Maas, Xie, Jurafsky, and Ng]{maas2015lexicon}
Andrew Maas, Ziang Xie, Dan Jurafsky, and Andrew Ng.
\newblock Lexicon-free conversational speech recognition with neural networks.
\newblock In \emph{Proceedings of the 2015 Conference of the North American
  Chapter of the Association for Computational Linguistics: Human Language
  Technologies}, pages 345--354, 2015.

\bibitem[Panayotov et~al.(2015)Panayotov, Chen, Povey, and
  Khudanpur]{panayotov2015librispeech}
Vassil Panayotov, Guoguo Chen, Daniel Povey, and Sanjeev Khudanpur.
\newblock Librispeech: an asr corpus based on public domain audio books.
\newblock In \emph{2015 IEEE International Conference on Acoustics, Speech and
  Signal Processing (ICASSP)}, pages 5206--5210. IEEE, 2015.

\bibitem[Park et~al.(2019)Park, Chan, Zhang, Chiu, Zoph, Cubuk, and
  Le]{park2019specaugment}
Daniel~S Park, William Chan, Yu~Zhang, Chung-Cheng Chiu, Barret Zoph, Ekin~D
  Cubuk, and Quoc~V Le.
\newblock Specaugment: A simple data augmentation method for automatic speech
  recognition.
\newblock In \emph{Interspeech}, 2019.

\bibitem[Prabhavalkar et~al.(2018)Prabhavalkar, Sainath, Wu, Nguyen, Chen,
  Chiu, and Kannan]{prabhavalkar2018minimum}
Rohit Prabhavalkar, Tara~N Sainath, Yonghui Wu, Patrick Nguyen, Zhifeng Chen,
  Chung-Cheng Chiu, and Anjuli Kannan.
\newblock Minimum word error rate training for attention-based
  sequence-to-sequence models.
\newblock In \emph{International Conference on Acoustics, Speech and Signal
  Processing (ICASSP)}, pages 4839--4843. IEEE, 2018.

\bibitem[Pratap et~al.(2018)Pratap, Hannun, Xu, Cai, Kahn, Synnaeve,
  Liptchinsky, and Collobert]{pratap2018}
Vineel Pratap, Awni Hannun, Qiantong Xu, Jeff Cai, Jacob Kahn, Gabriel
  Synnaeve, Vitaliy Liptchinsky, and Ronan Collobert.
\newblock wav2letter++: The fastest open-source speech recognition system.
\newblock \emph{arXiv preprint arXiv:1812.07625}, 2018.

\bibitem[Rao et~al.(2015)Rao, Peng, Sak, and Beaufays]{rao2015grapheme}
Kanishka Rao, Fuchun Peng, Ha{\c{s}}im Sak, and Fran{\c{c}}oise Beaufays.
\newblock Grapheme-to-phoneme conversion using long short-term memory recurrent
  neural networks.
\newblock In \emph{2015 IEEE International Conference on Acoustics, Speech and
  Signal Processing (ICASSP)}, pages 4225--4229. IEEE, 2015.

\bibitem[Santos and Zadrozny(2014)]{santos2014learning}
Cicero~D Santos and Bianca Zadrozny.
\newblock Learning character-level representations for part-of-speech tagging.
\newblock In \emph{Proceedings of the 31st International Conference on Machine
  Learning (ICML-14)}, pages 1818--1826, 2014.

\bibitem[Settle et~al.(2019)Settle, Audhkhasi, Livescu, and
  Picheny]{settle2019words}
Shane Settle, Kartik Audhkhasi, Karen Livescu, and Michael Picheny.
\newblock Acoustically grounded word embeddings for improved acoustics-to-word
  speech recognition.
\newblock In \emph{2015 IEEE International Conference on Acoustics, Speech and
  Signal Processing (ICASSP)}. IEEE, 2019.

\bibitem[Soltau et~al.(2016)Soltau, Liao, and Sak]{soltau2016neural}
Hagen Soltau, Hank Liao, and Hasim Sak.
\newblock Neural speech recognizer: Acoustic-to-word lstm model for large
  vocabulary speech recognition.
\newblock \emph{arXiv preprint arXiv:1610.09975}, 2016.

\bibitem[Sutskever et~al.(2014)Sutskever, Vinyals, and
  Le]{sutskever2014sequence}
Ilya Sutskever, Oriol Vinyals, and Quoc~V Le.
\newblock Sequence to sequence learning with neural networks.
\newblock In \emph{Advances in neural information processing systems (NIPS)},
  pages 3104--3112, 2014.

\bibitem[Synnaeve et~al.(2019)Synnaeve, Xu, Kahn, Grave, Likhomanenko, Pratap,
  Sriram, Liptchinsky, and Collobert]{synnaeve2019e2e}
Gabriel Synnaeve, Qiantong Xu, Jacob Kahn, Edouard Grave, Tatiana Likhomanenko,
  Vineel Pratap, Anuroop Sriram, Vitaliy Liptchinsky, and Ronan Collobert.
\newblock End-to-end asr: from supervised to semi-supervised learning with
  modern architectures.
\newblock \emph{arXiv preprint arXiv:1911.08460}, 2019.

\bibitem[Vaswani et~al.(2017)Vaswani, Shazeer, Parmar,
  et~al.]{vaswani2017attention}
Ashish Vaswani, Noam Shazeer, Niki Parmar, et~al.
\newblock Attention is all you need.
\newblock In \emph{Adv. NIPS}, 2017.

\bibitem[Zeyer et~al.(2018)Zeyer, Irie, Schl{\"u}ter, and
  Ney]{zeyer2018improved}
Albert Zeyer, Kazuki Irie, Ralf Schl{\"u}ter, and Hermann Ney.
\newblock Improved training of end-to-end attention models for speech
  recognition.
\newblock \emph{arXiv preprint arXiv:1805.03294}, 2018.

\end{thebibliography}
	\bibliographystyle{plainnat}
\end{small}

\end{document}